\documentclass{article}
\usepackage{ijcai17}
\usepackage{color}
\usepackage{times}
\usepackage[small]{caption}
\newcommand{\eat}[1]{}
\usepackage[table,xcdraw]{xcolor}
\usepackage{times}
\usepackage{epsfig}
\usepackage{graphicx}
\usepackage{epstopdf}

\usepackage{amsmath}
\usepackage{amssymb}
\usepackage{rotating}
\usepackage{array}
\usepackage{booktabs}
\usepackage{color}

\usepackage{longtable}

\title{Hierarchical LSTM with Adjusted Temporal Attention for Video Captioning}
\author{Jingkuan Song$^1$, Zhao Guo$^1$, Lianli Gao$^{1}$, Wu Liu$^{2}$, Dongxiang Zhang$^{1}$, Heng Tao Shen$^{1}$\\
	$^1$Center for Future Media and School of Computer Science and Engineering, \\University of Electronic Science and Technology of China, Chengdu 611731, China.\\ 
	$^2$Beijing University of Posts and Telecommunications, Beijing 100876, China.\\
	{jingkuan.song}@gmail.com, \{zhao.guo, lianli.gao, zhangdo\}@uestc.edu.cn,\\ liuwu@bupt.edu.cn, shenhengtao@hotmail.com}

\begin{document}

\maketitle

\begin{abstract}
Recent progress has been made in using attention based encoder-decoder framework for video captioning. However, most existing decoders apply the attention mechanism to every generated word including both visual words (e.g., "gun" and "shooting") and non-visual words (e.g. "the", "a").
However, these non-visual words can be easily predicted using natural language model without considering visual signals or attention.
Imposing attention mechanism on non-visual words could mislead and decrease the overall performance of video captioning.
To address this issue, we propose a hierarchical LSTM with adjusted temporal attention (hLSTMat) approach for video captioning. 
Specifically, the proposed framework utilizes the temporal attention for selecting specific frames to predict the related words, while the adjusted temporal attention is for deciding whether to depend on the visual information or the language context information. 
Also, a hierarchical LSTMs is designed to simultaneously consider both low-level visual information and high-level language context information to support the video caption generation. 
To demonstrate the effectiveness of our proposed framework, we test our method on two prevalent datasets: MSVD and MSR-VTT, and experimental results show that our approach outperforms the state-of-the-art methods on both two datasets.

\end{abstract}

\section{Introduction}

\begin{figure*}[ht]
\centering
\includegraphics[width=0.85\linewidth]{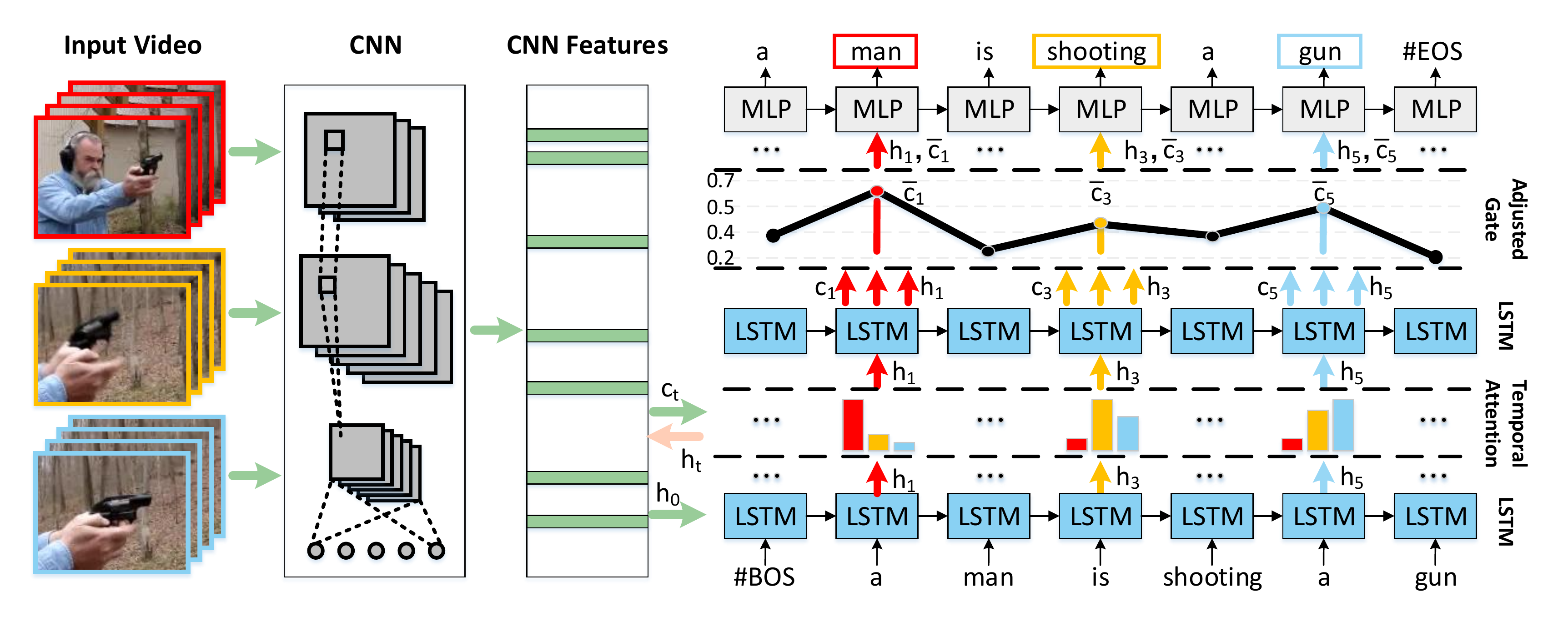}
\caption{The framework of our proposed method hLSTMat. To illustrate the effectiveness of our hLSTMat, each generated visual words (i.e., "man", "shooting" or "gun") is generated with visual information extracting from a set of specific frames. For instance, "man" is marked with "red", this indicates it is generated by using the frames marked with red bounding boxes, "shooting" is generated replying on the frames marked with "orange". Other non-visual words such as "a" and "is" are relying on the language model.}
\label{fig:framework}
\end{figure*}

Previously, visual content understanding~\cite{song2016optimized,GaoWSHSS17} and natural language processing (NLP) are not correlative with each other. Integrating visual content with natural language learning to generate descriptions for images, especially for videos, has been regarded as a challenging task. Video captioning is a critical step towards machine intelligence and many applications in daily scenarios, such as video retrieval \cite{wang2017survey,song2017quantization}, video understanding, blind navigation and automatic video subtitling. 

\eat{\textcolor{red}{Why use LSTM }}

Thanks to the rapid development of deep Convolutional Neural Network (CNN), recent works have made significant progress for image captioning \cite{vinyals2015show,xu2015show,Lu2016Knowing,karpathy2014deep,fang2015captions,chen2014learning,chen2016sca}. However, compared with image captioning, video captioning is more difficult due to the diverse sets of objects, scenes, actions, attributes and salient contents. Despite the difficulty there have been a few attempts for video description generation \cite{venugopalan2014translating,venugopalan2015sequence,yao2015describing,li2015summarization,45493}, which are mainly inspired by recent advances in translating with Long Short-Term Memory (LSTM). The LSTM is proposed to overcome the vanishing gradients problem by enabling the network to learn when to forget previous hidden states and when to update hidden states by integrating memory units. LSTM has been successfully adopted to several tasks, e.g., speech recognition, language translation and image captioning \cite{cho2015describing,venugopalan2014translating}. Thus, we follow this elegant recipe and choose to extend LSTM to generate the video sentence with semantic content.

Early attempts were proposed \cite{venugopalan2014translating,venugopalan2015sequence,yao2015describing,li2015summarization} to directly connect a visual convolution model to a deep LSTM networks. For example, Venugopalan \textit{et al.} \cite{venugopalan2014translating} translate videos to sentences by directly {concatenating} a deep neural network with a recurrent neural network.
More recently, attention mechanism \cite{GuLLL16} is a standard part of the deep learning toolkit, contributing to impressive results in neural machine translation \cite{luong2015effective}, visual captioning \cite{xu2015show,yao2015describing} and question answering \cite{Yang_2016_CVPR}. Visual attention models for video captioning make use of video frames at every time step, without explicitly considering the semantic attributes of the predicted words. For example, in Fig.~\ref{fig:framework}, some words (i.e., "man", "shooting" and "gun") belong to visual words which have corresponding canonical visual signals, while other words (i.e., "the", "a" and "is") are non-visual words, which require no visual information but language context information~\cite{Lu2016Knowing}. In other words, current visual attention models make use of visual information for generating each work, which is  unnecessary or even misleading. Ideally, video description not only requires modeling and integrating their {sequence dynamic temporal attention information into a natural language but also needs to take into account the relationship between sentence semantics and visual content} \cite{DBLP:journals/corr/GanGHPTGCD16}, which to our knowledge has not been simultaneously considered.

To tackle these issues, inspired by the attention mechanism for image captioning~\cite{Lu2016Knowing}, in this paper we propose a unified encoder-decoder framework (see Fig.~\ref{fig:framework}), named hLSTMat, a Hierarchical LSTMs with adjusted temporal attention model for video captioning. Specifically, first, in order to extract more meaningful spatial features, we adopt a deep neural network to extract a 2D CNN feature vector for each frame. Next, we integrate a hierarchical LSTMs consisting of two layers of LSTMs, temporal attention and adjusted temporal attention to decode visual information and language context information to support the generation of sentences for  videos description. Moreover, the proposed novel adjusted temporal attention mechanism automatically decides whether to rely on visual information or not. When relying on visual information, the model enforces the gradients from visual information to support video captioning, and decides where to attend. Otherwise, the model predicts the words using natural language model without considering visual signals. 

It is worthwhile to highlight the main contributions of this proposed approach:
1) We introduce a novel hLSTMat framework which automatically decides when and where to use video visual information, and when and how to adopt the language model to generate the next word for video captioning.
2) We propose a novel adjusted temporal attention mechanism which is based on temporal attention. Specifically, the temporal attention is used to decide where to look at visual information, while the adjusted temporal model is designed to decide when to make use of visual information and when to rely on language model.
A hierarchical LSTMs is designed to obtain low-level visual information and high-level language context information.
3) Experiments on two benchmark datasets demonstrate that our method outperforms the state-of-the-art methods in both BLEU and METEOR.

\section{The Proposed Approach}
\label{sec:methodology}
In this section, first we briefly describe how to directly use the basic Long Short-Term Memory (LSTM) as the decoder for video captioning task. Then we introduce our novel encoder-decoder framework, named hLSTMat (see Fig.~\ref{fig:framework}).

\subsection{A Basic LSTM for Video Captioning}

To date, modeling sequence data with Recurrent Neural Networks (RNNs) has shown great success in the process of machine translation, speech recognition, image and video captioning \cite{chen2014learning,fang2015captions,venugopalan2014translating,venugopalan2015sequence} etc.
Long Short-Term Memory (LSTM) is a variant of RNN to avoid the vanishing gradient problem \cite{bengio1994learning}.

\textbf{LSTM Unit.} A basic LSTM unit consists of three gates (input ${\mathbf{i}_t}$, forget ${\mathbf{f}_t}$ and output ${\mathbf{o}_t}$), a single memory cell ${\mathbf{m}_t}$. Specifically, ${\mathbf{i}_t}$ allows incoming signals to alter the state of the memory cell or block it. ${\mathbf{f}_t}$ controls what to be remembered or be forgotten by the cell, and somehow can avoid the gradient from vanishing or exploding when back propagating through time. Finally, ${\mathbf{o}_t}$ allows the state of the memory cell to have an effect on other neurons or prevent it. Basically, the memory cell and gates in a LSTM block are defined as follows:
\begin{equation}
\begin{aligned}
\mathbf{i}_{t} & = \sigma (\mathbf{W}_{i} \mathbf{y}_{t} +  \mathbf{U}_{i} h_{t-1} + \mathbf{b}_{i}) \\
\mathbf{f}_{t} & = \sigma (\mathbf{W}_{f} \mathbf{y}_{t} +  \mathbf{U}_{f} h_{t-1} + \mathbf{b}_{f}) \\
\mathbf{o}_{t} & = \sigma (\mathbf{W}_{o} \mathbf{y}_{t} +  \mathbf{U}_{o} h_{t-1} + \mathbf{b}_{o}) \\
\mathbf{g}_{t} & = \phi (\mathbf{W}_{g} \mathbf{y}_{t} +  \mathbf{U}_{g} h_{t-1} + \mathbf{b}_{g}) \\
\mathbf{m}_{t} & = \mathbf{f}_{t} \odot \mathbf{m}_{t-1}+ 
\mathbf{i}_{t} \odot \mathbf{g}_{t}         \\
\mathbf{h}_{t} & = \mathbf{o}_{t}\odot \phi(\mathbf{m}_{t})
\end{aligned}
\end{equation}
where the weight matrices $\mathbf{W}$, $\mathbf{U}$, and $\mathbf{b}$ are parameters to be learned. $\mathbf{y}_{t}$ represents the input vector for the LSTM unit at each time $t$. $\sigma$ represents the logistic sigmoid non-linear activation function mapping real numbers to $(0,1)$, and it can be thought as knobs that LSTM learns to selectively forget its memory or accept the current input. $\phi$ denotes the hyperbolic tangent function tanh. $\odot$ is the element-wise product with the gate value. For convenience, we denote $\mathbf{h}_{t},\mathbf{m}_{t}=\mathrm{LSTM}(\mathbf{y}_{t}, \mathbf{h}_{t-1}, \mathbf{m}_{t-1})$ as the computation function for updating the LSTM internal state.

\textbf{Video Captioning.} Given a video input $\mathbf{x}$, an encoder network $\phi_{E}$ encodes it into a continuous representation space:
\begin{equation}
\mathbf{V}=\{\mathbf{v}_{1}, \cdots, \mathbf{v}_{n}\} = \phi_{E}(\mathbf{x}).
\end{equation}
where $\phi_{E}$ usually denotes a CNN neural network, $n$ denotes the number of frames in $\mathbf{x}$, $\mathbf{v}_{i} \in \mathbb{R}^d$ is the frame-level feature of the $i$-th frame, and it is $d$-dimensional. Here, LSTM is chosen as a decoder network $\phi_{D}$ to model $\mathbf{V}$ to generate a description $\mathbf{z}=\{z_{1}, \cdots, z_{T}\}$ for $\mathbf{x}$, where $T$ is the description length. In addition, the LSTM unit updates its internal state $\mathbf{h}_{t}$ and the $t$-th word $z_{t}$ based on its previous internal state $\mathbf{h}_{t-1}$, the previous output $y_{t}$ and the representation $\mathbf{V}$:
\begin{equation*}
\left( \mathbf{h}_{t},z_{t} \right)
= \phi_{D}(\mathbf{h}_{t-1},y_{t},\mathbf{V})
\end{equation*}

In addition, the LSTM updates its internal state recursively until the end-of-sentence tag is generated. For simplicity, we named this simple method as basic-LSTM.

\subsection{Hierarchical LSTMs with Adjusted Temporal Attention for Video Captioning}
In this subsection, we introduce our hLSTMat framework, which consists of two components: 1) a CNN Encoder and 2) an attention based hierarchical LSTM decoder.

\subsubsection{CNN Encoders} The goal of an encoder is to compute feature vectors that are compact and representative and can capture the most related visual information for the decoder. Thanks to the rapid development of deep convolutional neural networks (CNNs), which have made a great success in large scale image recognition task \cite{he2015deep}, object detection \cite{ren2015faster} and visual captioning \cite{venugopalan2014translating}. High-level features can been extracted from upper or intermediate layers of a deep CNN network. Therefore, a set of well-tested CNN networks, such as the ResNet-152 model \cite{he2015deep} which has achieved the best performance in ImageNet Large Scale Visual Recognition Challenge, can be used as a candidate encoder for our framework. 

\begin{figure}[t]
	\centering
	\includegraphics[width=0.99\linewidth]{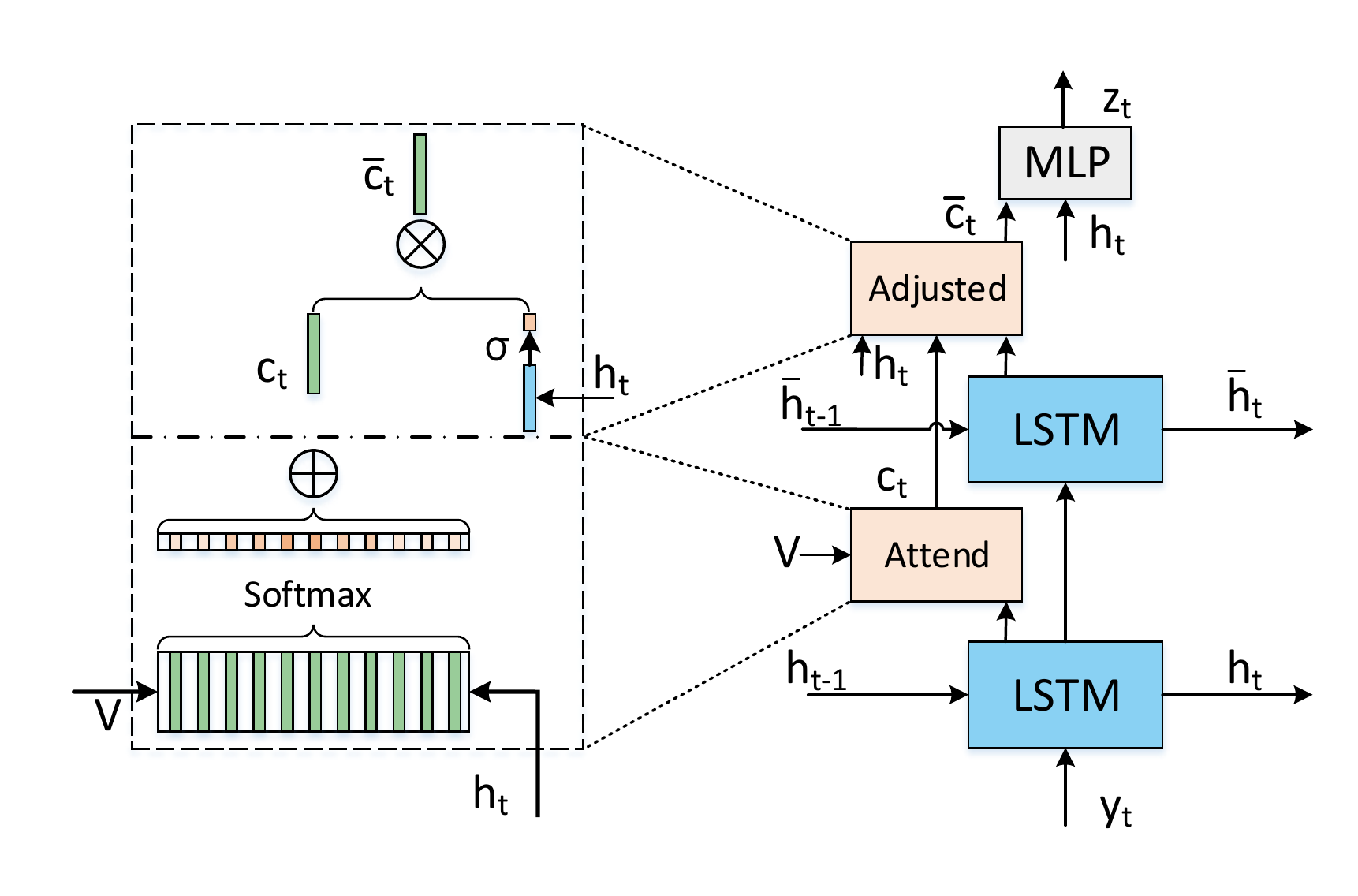}
	\caption{An illustration of the proposed method generating the t-th target word $z_t$ given a video.}
	\label{fig:our_attention}
\end{figure}

\subsubsection{Attention based Hierarchical Decoder}

Our decoder (see Fig.~\ref{fig:our_attention}) integrates two LSTMs. 
The bottom LSTM layer is used to efficiently decode visual features, and the top LSTM is focusing on mining deep language context information for video captioning.
We also incorporate two attention mechanisms into our framework. A temporal attention is to guide which frame to look, while the adjusted temporal attention is proposed to decide when to use visual information and when to use sentence context information.
The top MLP layer is to predict the probability distribution of each word in the vocabulary.

Unlike vanilla LSTM decoder, which performs mean pooling over 2D features across each video to form a fixed-dimension representation, attention based LSTM decoder is focusing on a subset of consecutive frames to form a fixed-dimensional representation at each time $t$. 
\begin{itemize}
\item{ Bottom LSTM Layer. For the bottom LSTM layer, the updated internal hidden state depends on the current word $y_{t}$, previous hidden state $\mathbf{h}_{t-1}$ and memory state $\mathbf{m}_{t-1}$:
\begin{equation}
\begin{aligned}
\mathbf{h}_{0},\mathbf{m}_{0} & =  \left[\mathbf{W^{ih}}; \mathbf{W^{ic}} \right]  Mean(\{\mathbf{v_{i}}\})  \\
\mathbf{h}_{t},\mathbf{m}_{t} &= LSTM(\mathbf{y}_{t}, \mathbf{h}_{t-1}, \mathbf{m}_{t-1}) \\
\end{aligned}
\end{equation}
where $\mathbf{y}_{t} = \mathbf{E}[y_{t}]$ denotes a word feature of a single word $y_{t}$. $Mean(\cdot)$ denotes a mean pooling of the given feature set $\mathbf{v_{i}}$. $\mathbf{W^{ih}}$ and $\mathbf{W^{ic}}$ are parameters that need to be learned.
}

\item{Top LSTM Layer. For the top LSTM, it takes the output of the bottom LSTM unit output $\mathbf{h}_{t}$, previous hidden state $\mathbf{\bar{h}}_{t-1}$ and the memory state $\mathbf{\bar{m}}_{t-1}$ as inputs to obtain the hidden state $\mathbf{\bar{h}}_{t}$ at time $t$, and it can be defined as below: 
\begin{equation}
\begin{aligned}
\mathbf{\bar{h}}_{t},\mathbf{\bar{m}}_{t}  & = LSTM( \mathbf{h}_{t}, \mathbf{\bar{h}}_{t-1}, \mathbf{\bar{m}}_{t-1}) \\
\end{aligned}
\end{equation}
}
\item{Attention Layers.
In addition, for attention based LSTM, context vector is in general an important factor, since it provides  meaningful visual evidence for caption generation \cite{yao2015describing}. In order to efficiently adjust the choose of visual information or sentence context information for caption generation, we defined an adjusted temporal context vector $\mathbf{\bar{c}}_{t} $ and a temporal context vector $\mathbf{c}_{t}$ at time $t$. See below: 
\begin{align}
\begin{aligned}
\mathbf{\bar{c}}_{t}   & = \psi(\mathbf{h}_{t}, \mathbf{\bar{h}}_{t}, \mathbf{c}_{t}),~~
\mathbf{c}_{t} & = \varphi(\mathbf{h}_{t}, \mathbf{V}) \\
\end{aligned}
\label{Eq:cv}
\end{align}
where $\psi$ denotes the function of our adjust gate, while $\varphi$ denotes the function of our temporal attention model. Moreover, $\mathbf{\bar{c}}_{t}$ denotes the final context vector through our adjusted gate, and $\mathbf{c}_{t}$ represents intermediate vectors calculated by our temporal attention model.
These two attention layers will be described in details in Sec.~\ref{sec.sub.tem.att} and Sec.~\ref{sec.sub.adj.tem.att}.
}

\item MLP layer. To output a symbol $z_{t}$, a probability distribution over a set of possible words is obtained using $\mathbf{h}_{t}$ and our adjusted temporal attention vector $\mathbf{\bar{c}}_{t} $:
\begin{equation}
\mathbf{p}_{t} = softmax \left( \mathbf{U}_{p} \phi(\mathbf{W}_{p}[\mathbf{h}_{t};\mathbf{\bar{c}}_{t}]+\mathbf{b}_{p}) + \mathbf{d}  \right)
\end{equation}
where $\mathbf{U}_{p}$, $\mathbf{W}_{p}$, $\mathbf{b}_{p}$ and $\mathbf{d}$ are parameters to be learned. Next, we can interpret the output of the softmax layer $\mathbf{p}_{t}$ as a probability distribution over words:
\begin{equation}
P(z_{t}|z_{<t}, \mathbf{V},\Theta)
\end{equation}
where $\mathbf{V}$ denotes the features of the corresponding input video, and $\Theta$ are model parameters.
\end{itemize}

To learn $\Theta$ in our modal, we minimize the negative logarithm of the likelihood:
\begin{equation}
min_{\Theta} - \sum_{t=1}^{T} log P(z_{t}|z_{<t}, \mathbf{V},\Theta)
\label{eq:loss}
\end{equation}
where $T$ denotes the total number of words in sentence. Therefore, Eq.\ref{eq:loss} is regarded as our loss function to optimize our model.

After the parameters are learned, we choose BeamSearch \cite{vinyals2015show} method to generate sentences for videos, which iteratively considers the set of the $k$ best sentences up to time $t$ as candidates to generate sentence of time $t+1$, and keeps only best k results of them. Finally, we approximates $D=argmax_{D^{'}}Pr(D^{'}|X)$ as our best generated description.
In our entire experiment, we set the beam size of BeamSearch as 5. 

\subsection{Temporal Attention Model}

\label{sec.sub.tem.att}

As mentioned above, context vector $\mathbf{c}_{t}$ is an important factor in encoder-decoder framework. To deal with the variability of the length of videos, a simple strategy \cite{venugopalan2014translating} is used to compute the average of features across a video, and this generated feature is used as input to the model at each time step:
\begin{align}
\mathbf{c}_{t}  = \frac{1}{n} \sum_{i=1}^{n} \mathbf{v}_{i}
\label{eq:meanpool1}
\end{align}

However, this strategy effectively collapses all frame-level features into a single vector, neglecting the inherent temporal structure and leading to the loss of information. Instead of using a simple average strategy (see Eq.~\ref{eq:meanpool1}), we wish to take the dynamic weight sum of the temporal feature vectors according to attention weights $\alpha_{t}^{i}$, which are calculated by a soft attention. For each $\mathbf{v}_{i}$ at $t$ time step, we use the follow function to calculate $\mathbf{c}_{t}$:
\begin{equation}
\mathbf{c}_{t}  = \frac{1}{n} \sum_{i=1}^{n} \alpha_{t}^{i} \mathbf{v}_{i}
\end{equation}
where at $t$ time step $\sum_{i=1}^{n} \alpha_{t}^{i} =1$.

In this paper, we integrate two LSTM layers, a novel temporal attention model for computing the context vector  $\mathbf{c}_{t}$ in Eq. \ref{Eq:cv} proposed in our framework. Given a set of video features $ \mathbf{V} $ and the current hidden state of the bottom layer LSTM $\mathbf{h}_{t}$, we feed them into a single neural network layer, and it returns an unnormalized relevant scores $\varepsilon_{t}$. Finally, a softmax function is applied to generate the attention distribute over the $n$ frames of the video:
\begin{equation}
\begin{aligned}
\varepsilon_{t} & =  \mathbf{w}^{T} tanh \left( \mathbf{W}_{a} \mathbf{h}_{t} + \mathbf{U}_{a} \mathbf{V} + \mathbf{b}_{a} \right) \\
\alpha_{t} & = softmax(\varepsilon_{t}) \\
\end{aligned}
\end{equation}
where $\mathbf{w}^{T}$, $\mathbf{W}_{a}$, $\mathbf{U}_{a}$ and $\mathbf{b}_{a}$ are parameters to be learned. $\alpha_{t} \in \mathbb{R}^{n}$ is the attention weight which quantifies the relevance of features in $\mathbf{V}$.

Different from \cite{yao2015describing}, we utilize the current hidden state instead of previous hidden state $\mathbf{h}_{t}$ generated by the first LSTM layer to obtain the context vector $\mathbf{c}_{t}$, which focuses on salient feature in the video.

\subsection{Adjusted Temporal Attention Model}
\label{sec.sub.adj.tem.att}
\eat{\textcolor{red}{Why adaptive and our adaptive}}


In this paper we propose an adjusted temporal attention model to compute a context vector $\mathbf{\bar{c}}_{t}$ in Eq.~\ref{Eq:cv}, shown in Fig. \ref{fig:our_attention}, to make sure that a decoder uses nearly no visual information from video frames to predict the non-visual words, and use the most related visual information to predict visual words. In our hierarchical LSTM network, the hidden state in the bottom LSTM layer is a latent representation of what the decoder already knows. With the hidden state $\mathbf{h}_{t}$, we extend our temporal attention model, and propose an adjusted model that is able to determine whether it needs to attend the video to predict the next word. In addition, a sigmoid function is applied to the hidden state $\mathbf{h}_{t}$ to further filter visual information.
\begin{equation}
\begin{aligned}
\mathbf{\bar{c}}_{t} & = \beta_{t} \mathbf{c}_{t} + (1-\beta_{t})\mathbf{\bar{h}}_{t} \\
\beta_{t} & = sigmoid(\mathbf{W}_s \mathbf{h}_{t})
\end{aligned}
\end{equation}
where $\mathbf{W}_s$ denotes the parameters to be learned and $\beta_{t}$ is adjusted gate at each time $t$. In our adjusted temporal attention model, $\beta_{t}$ is projected into the range of $[0, 1]$. When $\beta_{t}=1$, it indicates that full visual information is considered, while when $\beta_{t}=0$ it indicates that none visual information is considered to generate the next word.


\section{Experiments}
\label{sec:experimental}
We evaluate our algorithm on the task of video captioning. Specifically, we firstly study the influence of CNN encoders. Secondly, we explore the effectiveness of the proposed components. Next, we compare our results with the state-of-the-art methods.

\subsection{Datasets}
We consider two publicly available datasets that have been widely used in previous work.

\textbf{The Microsoft Video Description Corpus (MSVD).}  This video corpus consists of 1,970 short video clips, approximately 80,000 description pairs and about 16,000 vocabulary words~\cite{chen2011collecting}. Following \cite{yao2015describing,venugopalan2015sequence}, we split the dataset into training, validation and testing set with 1,200, 100 and 670 videos, respectively.

\textbf{MSR Video to Text (MSR-VTT).} In 2016, Xu \textit{et al.} \cite{xu2016msr} proposed a currently largest video benchmark for video understanding, and especially for video captioning. Specifically, this dataset contains 10,000 web video clips, and each clip is annotated with approximately 20 natural language sentences. In addition, it covers the most comprehensive categorizes (i.e., 20 categories) and a wide variety of visual content, and contains 200,000 clip-sentence pairs. 

\subsection{Implementation Details }
\subsubsection{Preprocessing} For MSVD dataset, we convert all descriptions to lower cases, and then use wordpunct\_tokenizer method from NLTK toolbox to tokenize sentences and remove punctuations. Therefore, it yields a vocabulary of 13,010 in size for the training split. For MSR-VTT dataset, captions have been tokenized, thus we directly split descriptions using blank space, thus it yields a vocabulary of 23,662 in size for training split.   
{Inspired by \cite{yao2015describing}, we preprocess each video clip by selecting equally-spaced 28 frames out of the first 360 frames and then feeding them into a CNN network proposed in \cite{he2015deep}. Thus, for each selected frame we obtain a 2,048-D feature vector, which are extracted from the $pool5$ layer.}
 
\subsubsection{Training details} In the training phase, in order to deal with sentences with arbitrary length, we add a begin-of-sentence tag $<$BOS$>$ to start each sentence and an end-of-sentence tag $<$EOS$>$ to end each sentence. In the testing phase, we input $<$BOS$>$ tag into our attention-based hierarchical LSTM to trigger video description generation process. For each word generation, we choose the word with the maximum probability and stop until we reach $<$EOS$>$.

In addition, all the LSTM unit sizes are set as 512 and the word embedding size is set as 512, empirically. Our objective function Eq.~\ref{eq:loss} is optimized over the whole training video-sentence pairs with mini-batch 64 in size of MSVD and 256 in size of MSR-VTT. We adopt adadelta \cite{zeiler2012adadelta}, which is an adaptive learning rate approach, to optimize our loss function. In addition, we utilize dropout regularization with the rate of 0.5 in all layers and clip gradients element wise at 10. We stop training our model until 500 epochs are reached, or until the evaluation metric does not improve on the validation set at the patience of 20.

\subsubsection{Evaluation metrics} To evaluate the performance, we employ two different standard evaluation metrics: BLUE \cite{papineni2002bleu} and METEOR \cite{banerjee2005meteor}. \eat{and CIDEr \cite{vedantam2015cider}.}

\subsection{The Effect of Different CNN Encoders}

\begin{table}[t]
\centering
\footnotesize
\caption{Experiment results on the MSVD dataset. We use different features to verify our hLSTMat method. }
\begin{tabular}{l|cccc|c}
\hline
Model            & B@1   & B@2   & B@3   & B@4   & METEOR  \\ \hline 
C3D            		  & 79.9  & 68.2  & 58.3  & 47.5  & 30.5  \\ 
GoogleNet             & 80.8  & 68.6  & 58.9  & 48.5  & 31.9  \\
Inception-v3          & 82.7  & 72.0  & 62.5  & 51.9  & 33.5  \\ 
ResNet-50			  & 80.9  & 69.1  & 59.5  & 49.0  & 32.3   \\
ResNet-101            & 82.2  & 70.9  & 61.4  & 50.8  & 32.7 \\
ResNet-152            & \textbf{82.9} & \textbf{72.2} & \textbf{63.0} & 
               		 \textbf{53.0} & \textbf{33.6}  \\ \hline
\end{tabular}
\label{tab.result_features}
\end{table}

\begin{table*}[t]
\centering
\footnotesize
\caption{The effect of different components and the comparison with the state-of-the-art methods on the MSVD dataset. The default encoder for all methods is ResNet-152.}
\begin{tabular}{l||c|c|c|c|c|c}
\hline
Model   & B@1   & B@2   & B@3   & B@4   & METEOR   & CIDEr \\ \hline 
basic LSTM            
		& 80.6     & 69.3     & 59.7     & 49.6     & 32.7 & 69.9  \\
MP-LSTM \cite{venugopalan2014translating}          
		& 81.1     & 70.2     & 61.0     & 50.4  & 32.5  & 71.0  \\ 
SA \cite{yao2015describing}
	    & 81.6     & 70.3     & 61.6     & 51.3  & 33.3  & 72.0 \\ \hline
basic+adjusted attention & 80.9 & 69.7 & 61.1& 50.2& 31.6&71.5\\   
hLSTMt
		& 82.5  & 71.9  & 62.0  & 52.1  & 33.3  & 73.5   \\
hLSTMat    & \textbf{82.9} & \textbf{72.2} & \textbf{63.0} & 
               		 \textbf{53.0} & \textbf{33.6} & \textbf{73.8} \\ \hline
\end{tabular}
\label{tab.result_methods}
\end{table*}

\begin{table*}[t]
\centering
\small
\caption{The performance comparison with the state-of-the-art methods on MSVD dataset. (V) denotes VGGNet, (O) denotes optical flow, (G) denotes GoogleNet, (C) denotes C3D and (R) denotes ResNet-152.}
\begin{tabular}{l||c|c|c|c|c|c}
\hline
Model            & B@1   & B@2   & B@3   & B@4   & METEOR  & CIDEr  \\ \hline
S2VT(V) \cite{venugopalan2015sequence}    
				 &  -    &  -    &  -    & -     & 29.2    & -   \\
S2VT(V+O)    
				 &  -    &  -    &  -    & -     & 29.8    & -   \\
HRNE(G) \cite{pan2015hierarchical}
	             & 78.4  & 66.1  & 55.1  & 43.6  & 32.1    & - \\
HRNE-SA (G)
		         & 79.2  & 66.3  & 55.1  & 43.8  & 33.1    & - \\
LSTM-E(V)\cite{pan2015jointly}
			     & 74.9  & 60.9  & 50.6  & 40.2  & 29.5    & - \\
LSTM-E(C)      & 75.7  & 62.3  & 52.0  & 41.7  & 29.9    & - \\
LSTM-E(V+C)        
				 & 78.8  & 66.0  & 55.4  & 45.3  & 31.0    & - \\
p-RNN(V) \cite{Yu_2016_CVPR}
			     & 77.3  & 64.5  & 54.6  & 44.3  & 31.1    & 62.1\\
p-RNN(C)       & 79.7  & 67.9  & 57.9  & 47.4  & 30.3  & 53.6 \\
p-RNN(V+C)  
				 & 81.5  & 70.4  & 60.4  & 49.9  & 32.6  & 65.8 \\ \hline
hLSTMt (R)
	 			& 82.5  & 71.9  & 62.0  & 52.1  & 33.3 & 73.5   \\
hLSTMat (R)
	             & \textbf{82.9} & \textbf{72.2} & 	 	  \textbf{63.0} & \textbf{53.0} & \textbf{33.6} &  \textbf{73.8} \\ \hline

\end{tabular}
\label{tab.result_msvd}
\end{table*}

To date, there are 6 widely used CNN encoders including C3D, GoogleNet, Inception-V3, ResNet-50, ResNet-101 and ResNet-152 to extract visual features. In this sub-experiment, we study the influence of different versions of CNN encoders on our framework. The experiments are conducted on the MSVD dataset, and the results are shown in Tab.~\ref{tab.result_features}. By observing Tab.~\ref{tab.result_features}, we find that by taking ResNet-152 as the visual decoder, our method performs best with 82.9\% B@1, 72.2\% B@2, 63.0\% B@3, 53.0\% B@4 and 33.6\% METEOR, while Inception-v3 is a strong competitor, with 82.7\% B@1, 72.0\% B@2, 62.5\% B@3, 51.9\% B@4 and 33.5\% METEOR. However, the gap between ResNet-152 and Inception-v3 is very small.

\subsection{Architecture Exploration and Comparison}

In this sub-experiment, we explore the impact of three proposed components, including basic LSTM proposed in Sec.2.1 (basic LSTM), hLSTMt which removes the adjusted mechanism from the hLSTMat, and hLSTMat, as well as comparing them with the state of the art methods: MP-LSTM \cite{venugopalan2014translating} and SA \cite{yao2015describing}. {In order to conduct a fair comparison, all the methods take ResNet-152 as the encoder.} We conduct the same experiments on the MSVD dataset. The experimental results are shown in Tab.~\ref{tab.result_methods}. It shows that our hLSTMat achieves the best results in all metrics with 82.9\% B@1, 72.2\% B@2, 63.0\% B@3, 53.0\% B@4 and 33.6\% METEOR. {Also, by comparing with SA which take previous hidden state to calculate temporal attention weight, our hLSTMt performs better for video captioning. Moreover, by comparing with hLSTMt, we find that adjusted attention mechanism can improve the performance of video captioning. We also add one-layer LSTM and adjusted attention as an additional baseline. Results show that the adjusted attention mechanism can improve the performance.}

\subsection{Compare with the-state-of-the-art Methods}


\subsubsection{Results on MSVD dataset} In this subsection, we show the comparison of our approach with the baselines on the MSVD dataset. Some of the above baselines only utilize video features generated by a single deep network, while others (i.e., S2VT, LSTM-E and p-RNN) make uses of both single network and multiple network generated features. Therefore, we first compare our method with approaches using static frame-level features extracted by a single network. In addition, we compare our method with methods utilized different deep features or their combinations.  The results are shown in Tab.\ref{tab.result_msvd}.
When using static frame-level features, we have the following observations: 
\\\noindent1) Compared with the best counterpart (i.e., p-RNN) which only takes spatial information, our method has 8.7\% improvement on B@4 and 2.5\% on METEOR.
\\\noindent2) The hierarchical structure in HRNE reduces the length of input flow and composites multiple consecutive input at a higher level, which increases the learning capability and enables the model encode richer temporal information of multiple granularities. Our approach (53.0\% B@4, 33.6\% METEOR) performs better than HRNE (43.6\% B@4, 32.1\% METEOR) and HRNE-SA (43.8\% B@4, 33.1\% METEOR). This shows the effectiveness of our model. 
	\eat{\item  S2VT uses one-layer LSTM as video encoder to explore videos' temporal information. Our approach achieves better result than S2VT (33.6\% vs 29.2\% on METERO).}
\\\noindent3) Our hLSTMat (53.0\% B@4, 33.6\% METEOR) can achieve better results than our hLSTMt (52.1\% B@4, 33.3\% METEOR). This indicates that it is beneficial to incorporate the adjusted temporal attention into our framework.

On the other hand, utilizing both spatial and temporal video information can enhance the video caption performance. VGGNet and GoogleNet are used to generate spatial information, while optical flow and C3D are used for capturing temporal information. For example, compared with LSTM-E(V) and LSTM-E (C), LSTM-E(V+C) achieves higher 45.3\% B@4 and 31.0\% METEOR. In addition, for p-RNN, p-RNN(V+C) (49.9\% B@4 and 32.6\% METEOR ) performs better than both p-RNN(V) (44.3\% B@4 and 31.3\% METEOR) and p-RNN(C) (47.4\% B@4 and 30.3\% METEOR).

Our approach achieves the best results (53.0\% B@4 and 33.6\% METEOR) using static frame-level features compared with approaches combining multiple deep features. For S2VT(V+O), LSTM-E(V+C) and p-RNN(V+C), they use two networks VGGNet/GoogleNet and optical flow/C3D to capture video's spatial and temporal information, respectively. Compared with them, our approach only utilizes ResNet-152 to capture frame-level features, which proves the effectiveness of our hierarchical LSTM with adjusted temporal attention model.

We adopt questionnaires collected from ten users with different academic backgrounds. Given a video caption, users are asked to score the following aspects: 1) Caption Accuracy, 2) Caption Information Coverage, 3) Overall Quality.
Results show that our method outperforms others at `Overall Quality', and `Caption Accuracy' with small margin. But it has lower value for `Information coverage' than p-RNN.

\begin{table}[]
\centering
\small
\caption{The performance comparison with the state-of-the-art methods on MSR-VTT dataset.}
\begin{tabular}{l||c|c}
\hline
Model                  		& B@4   & METEOR   \\  \hline
MP-LSTM (V)            & 34.8 & 24.8 \\
MP-LSTM (C)            	& 35.4 & 24.8 \\
MP-LSTM (V+C)        & 35.8 & 25.3 \\
SA (V)       	        & 35.6 & 25.4 \\
SA (C)                    & 36.1 & 25.7 \\
SA (V+C)   	    	& 36.6 & 25.9 \\ \hline
hLSTMt (R)		            & \textbf{37.4} & \textbf{26.1} \\ 
hLSTMat (R)    		            & \textbf{38.3} & \textbf{26.3} \\ \hline
\end{tabular}
\label{tab.result_msr}
\end{table}

\subsubsection{Results on MSR-VTT dataset} We compare our model with the state-of-the-art methods on the MSR-VTT dataset, and the results are shown in Tab.~\ref{tab.result_msr}. 
Our model performs the best on all metrics, with 38.3\% @B4 and 26.3\% METEOR. Compared with our methods using only temporal attention, the performance is improved by 1.1\% for @B4, and 0.2\% for METEOR. This verifies the effectiveness of our method.

\eat{
 \textcolor{red}{Add the references in this section and Tab \ref{rsmsvd}}
 
To evaluate the performance of our proposed hLSTMat algorithm, we compare it with several state-of-the-art methods: basic encoder-decoder model (S2VT) ~\cite{DBLP:conf/iccv/VenugopalanRDMD15}, soft attention based LSTM network (SA) ~\cite{DBLP:conf/iccv/YaoTCBPLC15}, hierarchical processing based decoders (HRNE-VC) ~\cite{DBLP:conf/cvpr/PanXYWZ16} and p-RNN-VC ~\cite{DBLP:conf/cvpr/YuWHYX16} and deep convolutional GRU based framework (GRU-RCN) ~\cite{DBLP:journals/corr/BallasYPC15}. Note that all the compared methods use multiple modality visual features, while our approach only takes the ResNet-152 feature as input.

The experiments are conducted on both MSVD and MSR-VTT datasets, and the results are shown in Tab.Tab \ref{rsmsvd} and Tab.Tab \ref{tab.result_msr}, respectively. From both tables, we can see that our hLSTMat outperforms the existing approaches and achieves the best results on both datasets.  

}

\eat{
\section*{Acknowledgments}

}

\section{Conclusion and Future Work}
In this paper, we introduce a novel hLSTMat encoder-decoder framework, which integrates a hierarchical LSTMs, temporal attention and adjusted temporal attention to automatically decide when to make good use of visual information or when to utilize sentence context information, as well as to simultaneously considering both low-level video visual features and language context information. Experiments show that hLSTMat achieves state-of-the-art performances on both MSVD and MSR-VTT datasets. 
In the future, we consider incorporating our method with both temporal and visual features to test the performance.

\section*{Acknowledgments}
This work is supported by the Fundamental Research Funds for the Central Universities (Grant No. ZYGX2014J063, No. ZYGX2014Z007) and the National Natural Science Foundation of China (Grant No. 61502080, No. 61632007, No. 61602049).

{\small
\bibliographystyle{named}

}

\end{document}